\RequirePackage{fix-cm}
\documentclass[twocolumn]{svjour3}          
\smartqed  

\usepackage{cite}
\usepackage[pdftex]{graphicx}
\usepackage{ragged2e}
\usepackage[tight,footnotesize]{subfigure}
\usepackage{rotating}
\usepackage{graphicx}
\usepackage{amsmath,amssymb} 
\usepackage{array}
\usepackage{multirow}
\usepackage{colortbl}
\usepackage{amsfonts}
\usepackage{pifont}
\usepackage{xspace}
\usepackage{etoolbox}
\usepackage{overpic}
\usepackage{color}
\usepackage{microtype}

\usepackage{esvect}		
\usepackage{algorithm}
\usepackage{algorithmic}
\usepackage[breaklinks=true]{hyperref}

\newcommand{\figref}[1]{Fig. \ref{#1}}
\newcommand{\tabref}[1]{Tab. \ref{#1}}
\newcommand{\equref}[1]{Eq.~(\ref{#1})}

\newcommand{\algref}[1]{Alg. \ref{#1}}

\newcommand{\orcid}[1]{\href{https://orcid.org/#1}{\includegraphics[width=10pt]{ocrid.png}}}

\def\ie{\emph{i.e.}}

\def\etal{{\em et al}}

\usepackage{hyperref}
\hypersetup{breaklinks=true,citecolor=blue, colorlinks}

\graphicspath{{Imgs/}}
\DeclareGraphicsExtensions{.pdf,.jpg,.png}

\usepackage{silence}
\journalname{Research Article}

\begin{document}

\title{Implicit Non-Causal Factors are Out via Dataset Splitting for Domain Generalization Object Detection}


\author{ Zhilong Zhang\textsuperscript{1,2} \and Lei Zhang\textsuperscript{1,2*} \and Qing He\textsuperscript{1,2} \and Shuyin Xia\textsuperscript{3} \and Guoyin Wang\textsuperscript{4} \and Fuxiang Huang\textsuperscript{1,5}  
}

\authorrunning{Z. Zhang \etal} 

\institute{
1. Chongqing Key Laboratory of Bio-perception and Multimodal Intelligent Information Processing, Chongqing University, Chongqing, China. (E-mail: zhangzhilong@stu.cqu.edu.cn, \{leizhang, qinghe\}@cqu.edu.cn).\\
2. School of Microelectronics and Communication Engineering, Chongqing University, Chongqing 400044, China. \\
3. Chongqing Key Laboratory of Computational Intelligence, Key Laboratory of Cyberspace Big Data Intelligent Security, Ministry of Education,  Chongqing University of Posts and Telecommunications, Chongqing 400065, China. (E-mail: xiasy@cqupt.edu.cn).\\
4. National Center for Applied Mathematics in Chongqing, Chongqing Normal University, Chongqing 401331, China. (E-mail: wanggy@cqnu.edu.cn).\\
5. School of Data Science, Lingnan University, Hong Kong, China. (E-mail: fxhuang1995@gmail.com).\\
\textsuperscript{*}\textit{Corresponding author: Lei Zhang (leizhang@cqu.edu.cn)}.
}

\date{Received: date / Accepted: date}

\maketitle
\begin{abstract}
{Open world object detection faces a significant challenge in domain-invariant representation, i.e., implicit non-causal factors. 
Most domain generalization (DG) methods based on domain adversarial learning (DAL) pay much attention to learn domain-invariant information, but often overlook the potential non-causal factors.
We unveil two critical causes: 1) The domain discriminator-based DAL method is subject to the extremely sparse domain label, i.e., assigning only one domain label to each dataset, thus can only associate explicit non-causal factor, which is incredibly limited.
2) The non-causal factors, induced by unidentified data bias, are excessively implicit and cannot be solely discerned by conventional DAL paradigm.
Based on these key findings, inspired by the Granular-Ball perspective, we propose an improved DAL method, i.e., GB-DAL. The proposed GB-DAL utilizes Prototype-based Granular Ball Splitting (PGBS) module to generate more dense domains from limited datasets, akin to more fine-grained granular balls, indicating more potential non-causal factors. 
Inspired by adversarial perturbations akin to non-causal factors, we propose a Simulated Non-causal Factors (SNF) module as a means of data augmentation to reduce the implicitness of non-causal factors, and facilitate the training of GB-DAL. 
Comparative experiments on numerous benchmarks demonstrate that our method achieves better generalization performance in novel circumstances.}


\end{abstract}

\section{Introduction}
\label{sec:intr}

Deep neural networks (DNNs) have demonstrated excellent performance in various computer vision tasks, such as image classification and object detection \cite{ren2015faster,redmon2016you,liu2016ssd,lin2017focal,deng2020global,wang2021harmonic}. However, the domain shift \cite{5995347} has limited the application of DNNs in the open world. To alleviate this problem, unsupervised domain adaptation (DA) \cite{chen2018domain, he2019maf, hsu2020every, he2020atf,huang2024gh} learns domain-invariant representations by aligning the distributions of target and source domains through domain adversarial learning (DAL) techniques. Nevertheless, in open-world scenarios, the target domain is often unseen. 

Domain generalization (DG) \cite{mansilla2021domain,li2018learning,arjovsky2019invariant,huang2021fsdr,xu2021fourier,qiao2020learning,zhang2023rand} goes a step further, aiming to train a model on single or multiple source domains that can generalize to unknown target domains.
Existing DG models are devoted to mining domain-invariant features, but neglect a key fact that many potential non-causal factors are still implicitly embedded in domain-invariant representations, resulting in label spurious correlation, a hazard to accuracy.

\begin{figure}[tb]
	\centering
	\subfigure[Dataset A\label{fig:data_bias_a}]{
		\includegraphics[width=0.48\linewidth]{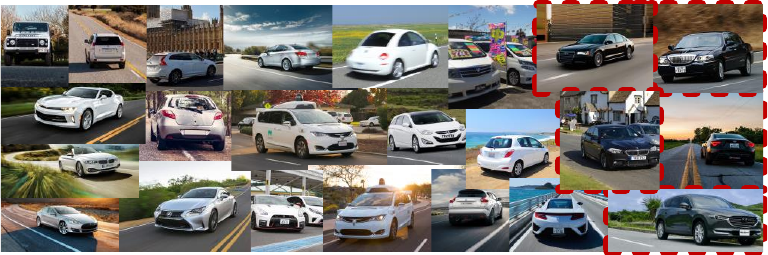}}
	\subfigure[Dataset B\label{fig:data_bias_b}]{
		\includegraphics[width=0.48\linewidth]{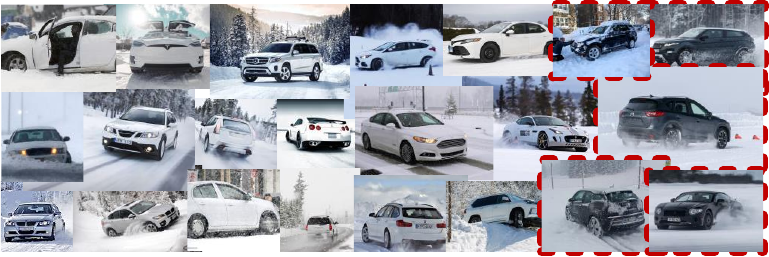}}	
	\caption{ Visual representations of inter- and intra-dataset non-causal factors.
		Road condition that differ in the two datasets are regarded as \textbf{inter-dataset non-causal factors}, while vehicle color that reflect data bias within the dataset are considered as \textbf{intra-dataset non-causal factors}.		
	}
	\label{fig:data_bias}
\end{figure}

We explore and unveil two essential causes why non-causal factors cannot be filtered in existing DG methods based on domain adversarial learning (DAL): 
1) \textit{Sparsity of domain labels.} 
Most DAL based methods \cite{ganin2015unsupervised,lin2021domain} treat each dataset as an individual domain and allocate only a single domain label to each dataset. Under such a sparse domain labeling scheme, DAL primarily focuses on filtering out explicit non-causal factors across datasets, but often overlooks the implicit non-causal factors within individual datasets.
As illustrated in \figref{fig:data_bias}, the two datasets exhibit significant differences in road conditions, with one dataset featuring snow-covered surfaces while the other dry surfaces. Additionally, an imbalance in vehicle color distribution is also observed within individual dataset, where white vehicles dominate while black vehicles constitute only a small proportion.
Under sparse domain label supervision at the dataset level, the inter-dataset non-causal factor (road conditions) can be easily focused and filtered out by DAL. 
However, within the dataset, the intra-dataset non-causal factor (vehicle color), induced by data bias, is often overlooked, resulting in learned domain-invariant features still encompass non-causal factors.
This issue can be alleviated by introducing more fine-grained and denser domain labels, which is one of the objectives of this paper. 
2) \textit{Implicitness of non-causal factors.}
In the scenario of extreme data bias where the dataset exclusively comprises white cars without any other color variations, the non-causal factor (vehicle color) may be erroneously deeply embedded into the discriminative features of car, making it too implicit to be recognized by DAL.
Therefore, to expose such non-causal factors, one can achieve this by augmenting additional non-causal elements to compensate for data bias.

Based on the above key findings and challenges, this paper is dedicated to answering two critical questions: 
1) \textit{How to obtain denser domain labels for fine-grained domain adversarial learning to unveil more non-causal factors?} 2) \textit{How to explicitly augment non-causal factors to bridge data bias and facilitate training?}

%

\begin{figure}[tb]
	\centering
	\subfigure[sparse domain label\label{fig:unsupervised_cluster_a}]{
		\includegraphics[width=0.45\linewidth]{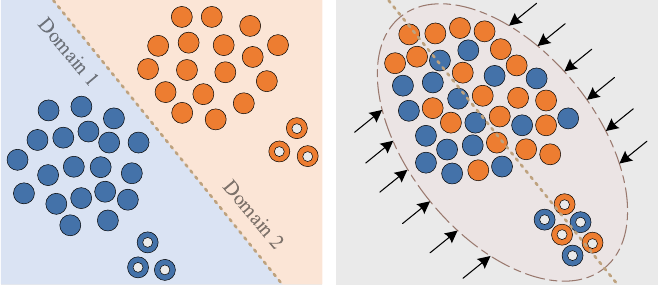}}\quad
	\subfigure[dense domain label\label{fig:unsupervised_cluster_b}]{
		\includegraphics[width=0.45\linewidth]{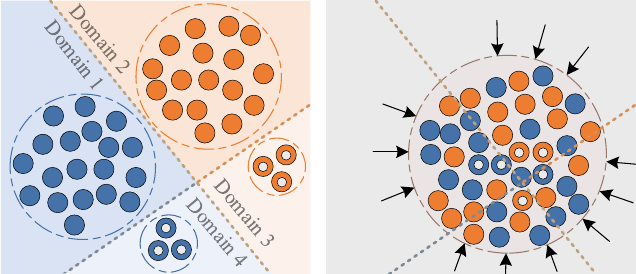}}

	\caption{
    Comparison of the alignment effects of DAL based on sparse and dense domain labels. (a) With sparse domain labels (\ie, dataset-level), DAL eliminates the inter-dataset non-causal factors (represented by color), but ignoring intra-dataset non-causal factors (represented by filling pattern). (b) With denser domain labels (\ie, datasets splits), intra-dataset non-causal factors can also be filtered out.
	}
	\label{fig:unsupervised_cluster}
\end{figure}

\begin{figure}[tb] 
	\centering
	\subfigure[Proportion of non-causal factors in each domain\label{fig:Distr}]{
		\includegraphics[width=0.47\linewidth]{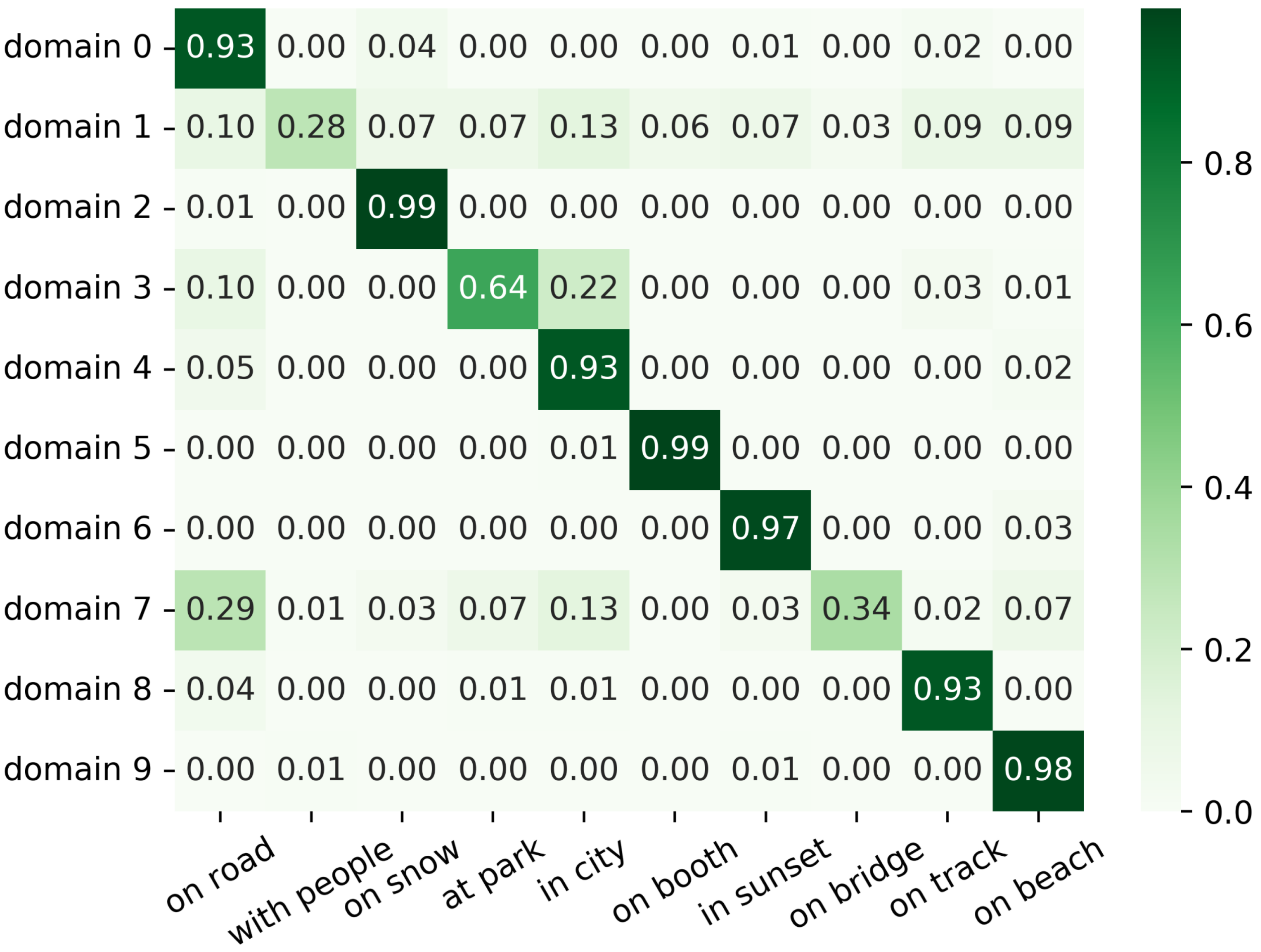}}\quad
	\subfigure[Visualization of different domains\label{fig:Visual}]{
		\includegraphics[width=0.39\linewidth]{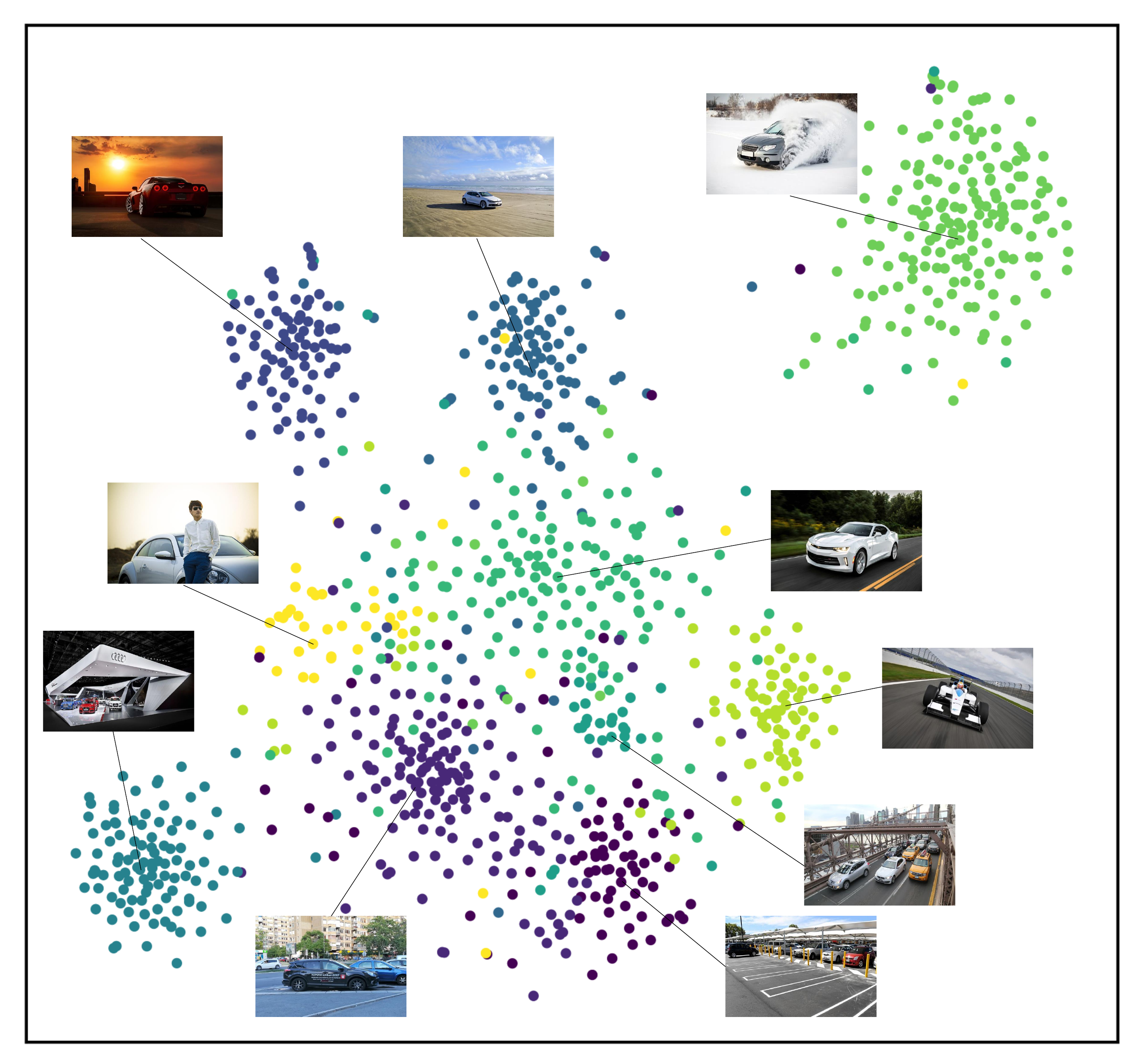}}
	\caption{Results of the domain label assignment of the PGBS module on cars under spurious correlations in NICO \cite{he2021towards}.
    }

	\label{fig:visual}
\end{figure}

For the $1^{st}$ question, inspired by Granular-Ball computing \cite{quadir2024granular,zhang2023incremental}, we propose a granular-ball based domain adversarial learning (GB-DAL) method, which uses a prototype-based granular ball splitting (PGBS) module to furnish more fine-grained domain labels from limited datasets for DAL. \figref{fig:unsupervised_cluster} intuitively shows that, compared with sparse domain labels (a), DAL with dense domain labels (b) can align features from more diverse directions, thereby filtering out a broader spectrum of non-causal factors.
Furthermore, we conduct a preliminary experiment to validate the capability of the PGBS module in capturing diverse non-causal factors through the split domain labels. We employ the PGBS module to assign domain labels to cars in the Out-of-Distribution (OOD) dataset i.e., NICO\cite{he2021towards}, that exhibits varying spurious correlation factors (e.g., on road, with people). As illustrated in \figref{fig:visual}, by applying PGBS to vehicles characterized by multiple non-causal factors, its capability in unveiling non-causal factors via domain label assignment is demonstrated. Specifically, as shown in \figref{fig:visual} (a), each domain identified by the PGBS module predominantly contains vehicles sharing a specific non-causal factor.  This indicates that PGBS can effectively disentangle different types of non-causal factors.  These domain distinctions then serve as a foundation for subsequent domain adversarial learning to eliminate such non-causal factors, ultimately enhancing generalization across domains. Additionally, \figref{fig:visual} (b) provides the visualization of different domains, which further indicates the ability of PGBS in segmenting data with distinct non-causal factors, thereby supporting improved generalization across domains.

Regarding the $2^{nd}$ question, addressing how to effectively augment non-causal factors, we approach it from the essence of non-causal factors. Although non-causal factors lack intrinsic causal relationship with the class labels, they are erroneously associated with class discrimination by the model. Based on this property, we observe that non-causal factors and adversarial perturbations share two critical properties that motivate our design. On one hand, as an interference factor, adversarial perturbations also lack a causal relationship with class labels, yet they can significantly affect the model's classification decision. On the other hand, adversarial perturbations are human imperceptible, while non-causal factors typically do not cause recognition difficulties for the human eye. This indicates that adversarial perturbations, in essence, can also be regarded as a type of non-causal factor. Motivated by this insight, we propose to Simulate Non-causal Factors (SNF) by employing adversarial perturbations to augment non-causal factors and mitigate data bias.

The main contributions of this paper are summarized as follows:
\begin{itemize}
\item We reveal two key insights: 1) existing DAL-based methods struggle to adequately identify and handle non-causal factors due to the sparsity of domain labels, and 
2) The non-causal factors induced by data bias are too implicitly to be effectively filtered out solely by DAL. 

\item We propose a granular-ball based fine-grained domain adversarial learning (GB-DAL) method via datasets split to filter both explicit (inter-dataset) and implicit (intra-dataset) non-causal factors.

\item We unveil the similarity between adversarial perturbations and non-causal factors, and propose a Simulated Non-causal Factors (SNF) module as a means of data augmentation to compensate for data bias and facilitate model training.
\end{itemize}

\section{Related Work}
\label{sec:relatedwork}

\textbf{Domain Adaptive Object Detection (DAOD)} aims to enhance the generalization capability of object detectors on the target domain by employing domain adaptation strategies to train the detectors on both the label-rich source domain and the unlabeled target domain. 
DAOD methods can be primarily categorized into two major classes: adversarial learning-based approaches and image reconstruction-based methods. Among the adversarial learning-based approaches, \cite{chen2018domain} pioneered the introduction of DAL into the DAOD task, achieving domain adaptation through feature space alignment between the source and target domains. Building upon this foundation, \cite{saito2019strong} proposed a local domain classifier network, which enables stronger feature alignment at the patch level. Regarding image reconstruction-based methods, \cite{arruda2019cross} was the first to employ the CycleGAN framework \cite{zhu2017unpaired}, generating target domain-style samples through image-to-image translation, thereby reducing the domain discrepancy.
However, on the one hand, the requirement of target domain data limits the practicability of DA. On the other hand, most DA methods relying on DAL cannot mine and filter the potential non-causal factors implied in domain invariant features due to the sparsity of dataset-level domain labels.

\textbf{Domain Generalization (DG)} aims to learn a model from one or multiple source domains that can generalize to unseen target domains without requiring access to target domain data.
Existing DG methods can be categorized into three main approaches: domain-invariant learning, domain augmentation, and other learning strategies.
First, the domain-invariant learning strategy aims to learn domain invariant representations from source domains.
\cite{li2018domain} proposed a novel framework based on Adversarial Autoencoders (AAE), which achieves DG through adversarial feature learning. 
Second, domain augmentation methods aim to increase the diversity of source domains and facilitate domain invariant representations.
\cite{xu2021fourier} mixes the amplitude spectrum of two images to augment source domains.
\cite{qiao2020learning} proposed an adversarial domain augmentation approach to enhance the DG in a single-source domain setting.
Third, there are learning strategies like \cite{li2018learning} which firstly adopts meta-learning for DG, following the idea of learning to learn domain-invariant components from different domains. Recently, \cite{xu2023mad} propose a multi-view domain discriminator based DG model for object detection and aim to filter non-causal factors, but it is still modeled on the conventional DAL paradigm with dataset-level domains.
Most existing DG methods are unable to filter potential non-causal factors due to the sparsity of domain labels in DAL and implicitness of non-causal factors.
We wish to relax the domain discriminator with dense but free domain labels to mine and filter more potential non-causal factors.

\section{Methods}
\label{sec:methods}
\subsection{Overview}
\begin{figure}[tb]
	\centering
	\includegraphics[width=1.0\linewidth]{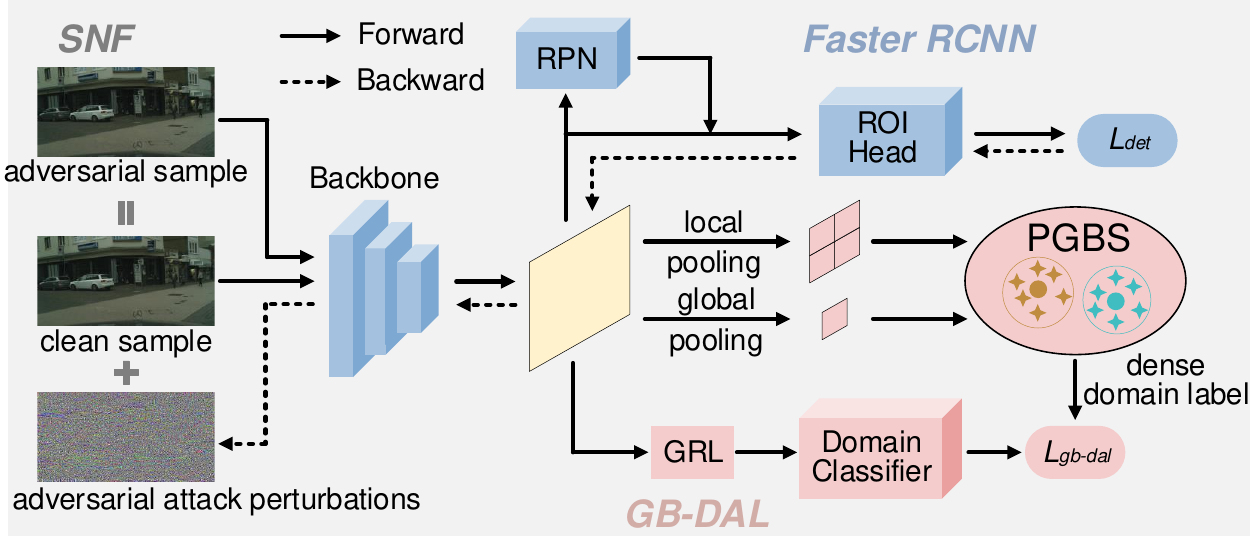}
	\caption{
		Diagram of the proposed domain generalized object detector, featuring two novel components: GB-DAL and SNF. SNF employs adversarial perturbations to simulate and expose hidden non-causal factors. In GB-DAL, the PGBS module (Prototype-based Granular Ball Splitting) computes dense domain labels via datasets split for domain adversarial learning, facilitating the identification and filtering of additional non-causal factors. GRL means the Gradient Reversal Layer.
	}
	\label{fig:overview}
\end{figure}

Existing DG methods are devoted to learning domain invariant features but overlook the potential domain-agnostic non-causal factors, which are detrimental to DG due to label spurious correlation. As illustrated in  \figref{fig:overview}, the proposed framework comprises two inherently integrated components, GB-DAL and SNF, developed from a unified causal learning perspective. The GB-DAL employs a PGBS module to generate dense domain labels, facilitating the systematic identification and filtering of intra-dataset non-causal factors through domain adversarial learning. Meanwhile, the SNF module serves as a causally motivated intervention mechanism, generating adversarial perturbations to explicitly simulate non-causal factors. This targeted intervention reveals spurious correlations during training and provides essential adversarial examples, enhancing GB-DAL's ability to learn robust causal feature representations. 

\textbf{Problem definition.} We make the following definitions to formalize the DG problem.
The source domain is denoted as $D_s=\{X_s, Y_s\}$.
The feature extractor $F(\cdot)$ can extract feature $f=F(x_s)$ from input images $x_s$.
Feature $f$ contains causal and non-causal factors $\{f_{cau}, f_{non}\}$, and the non-causal factors can be further divided into two parts $\{f_{non}^{inter}, f_{non}^{intra}\}$, in which $f_{non}^{inter}$ denotes inter-domain non-causal factors and $f_{non}^{intra}$ denotes intra-domain non-causal factors.
As discussed before, the domain-invariant features $f_{inv}$ learned by existing DG models can exclude inter-domain non-causal factors $f_{non}^{inter}$, but wrongly include potential intra-domain non-causal factors $f_{non}^{intra}$.

\subsection{The Proposed GB-DAL}
\subsubsection{Background and Preliminaries} 
Domain adversarial learning (DAL) \cite{ganin2015unsupervised} has emerged as a prominent technique in domain generalization, aiming to extract shared features across different domains.  By minimizing the $\mathcal{A}$-Distance between features from various domains, DAL seeks to enhance the model's robustness to domain shifts. From an implementation perspective, DAL can be formulated as a minimax optimization problem between feature extractor $\mathcal{F}$ and the ideal domain discriminator $h$:
\begin{equation}
	\begin{split}
		\min_{\mathcal{F}}d_{\mathcal{A}}(D_{s1},D_{s2})=&\underbrace{\max_{\mathcal{F}}\min_{h \in \mathcal{H}}err(h(f),y_{d})}_{Standard \ DAL} \\
		\Rightarrow&\underbrace{\max_{\mathcal{F}}\min_{h \in \mathcal{H}}err(h(f),y_{gb})}_{Our \ GB-DAL},
	\end{split}
	\label{eq:1}
\end{equation}
where $\mathcal{H}$ denotes a hypothesis set of all possible domain discriminators, $h(\cdot)$ is one of the domain discriminators in $\mathcal{H}$, $y_{d}$ denotes human-defined domain labels, $y_{gb}$ denotes domain labels obtained by the PGBS module.
However, due to the extreme sparsity of manually defined domain labels $y_{d}$, domain discriminator $h(\cdot)$ supervised by them can only capture the inter-domain non-causal factors $f_{non}^{inter}$ while neglecting other intra-domain non-causal factors $f_{non}^{intra}$. 

\subsubsection{Granular Ball Perspective}
Granular-ball computing (GB) ~\cite{xia2023granular,cao2024open} is a clustering concept inspired by human's multi-granularity cognition. Unlike traditional point- and pixel-based methods, GB represents data as hyper-spherical granules (granular balls) with adaptive size and density. It has been successfully applied to classification and clustering tasks, and improves robustness by filtering out noise at coarse granularities while preserving fine-grained structures. In accordance with previous research \cite{xia2023granular,cao2024open}, the multi-granularity granular-ball computing model can be formally articulated as follows:
\begin{equation}
	\begin{split}
		f(x,\vv{\alpha})&\longrightarrow g(G B,\vv{\beta}) \\
		s.t.\quad\mathrm{min}\quad&\frac{n}{\sum_{j=1}^{k}|G B_{j}|}+ K  \\
		s.t.\ qua&lity(G B_{j})\geq T,
	\end{split}
	\label{eq:granular-ball}
\end{equation}
where \( f(x, \vv{\alpha}) \) represents the existing learning model, with sample \( x \) as the input and \( \vv{\alpha} \) as the model parameters. \( g(GB, \vv{\beta}) \) represents the granular ball computing learning model, with the granular ball \( GB \) as the input and \( \vv{\beta} \) as the model parameters. $GB_j$ represents a granular-ball generated on the dataset $D=\{x_{i},i= 1,2,...,n\}$, \( n \) denotes the total number of samples in dataset \( D \), and \( K \) represents the total number of granular balls generated on \( D \), and \( | \cdot | \) indicates the number of samples within a set. \(quality()\) measures the quality of granular balls using loss functions based on metric learning, with $T$ as the quality threshold.


In general, DAL assigns a single domain label to each dataset, analogous to utilizing only one coarse-grained granular-ball to represent the entire dataset (\ie, $k$ is set to 1 in \equref{eq:granular-ball}). This results in that the domain-invariant features learned by conventional DAL still encompass numerous intra-dataset non-causal factors. Thus, we advocate multiple fine-grained granular-balls (with a larger value of $K$) to represent a dataset (\ie, denser fine-grained domains via dataset splitting). This, to a large extent, allows DAL to focus on intra-dataset non-causal factors and purify them in domain-invariant features, facilitating causality-invariant feature learning. The Granular-ball based dataset splitting technique named PGBS is introduced in Section \ref{sect33}.

\subsubsection{Cross-domain Object Detector with GB-DAL}
As depicted in \figref{fig:overview}, Faster RCNN serves as the underlying detector in our approach. Given an image $x$, processed through the feature extractor $F(\cdot)$ to obtain the feature map $f$, on one hand, the feature map $f$ is fed into the RPN and subsequent ROI Head for classification and regression; on the other hand, it is input into GB-DAL for fine-grained domain alignment. 
More specifically, in GB-DAL, the feature map $f^i$ (extracted from the $i$-th training image) undergoes average pooling layer $P(\cdot)$ to obtain $N$ pooled features $f^{i,j}_{pool}$, where $j \in \{1,2,\dots,N\}$. Subsequently, the PGBS module is employed to derive domain labels $y^{i,j}_{gb}$ for each pooled feature $f^{i,j}_{pool}$. Concurrently, the feature map $f^i$ passes through a gradient reverse layer (GRL) \cite{ganin2015unsupervised} and a domain classifier composed of multiple convolutional layers to obtain pixel-level domain predictions $p^{i,(u,v)}$. 
Finally, minimizing the cross-entropy loss between $p^{i,(u,v)}$ and $y^{i,j}_{gb}$ is adopted as the optimization objective for the domain classifier. It is noteworthy that, due to the utilization of GRL, the optimization objective for the feature extractor is to maximize this loss.

We perform GB-DAL based alignment at two levels: global features and local features of the image. 

\textbf{Local GB-DAL:} We obtain the local features $f^{i,j}_{pool}, j \in \{1,2,\dots,N\}$ of the image by performing adaptive average pooling on the feature map $f^i$. Then, the PGBS module is used to generate fine-grained domain label $y^{i,j}_{gb}$ for each local feature $f^{i,j}_{pool}$. 
These domain labels are subsequently utilized to supervise the domain classifier, facilitating the exploration of additional non-causal factors. The optimization loss for the local GB-DAL can be formally defined as follows:
\begin{equation}
	{\mathcal{L}}_{local}=-\sum_{i,u,v}y^{i,j}_{gb}\log{p^{i,{(u,v)}}} \quad \text{if} \: f^{i,(u,v)} \in f^{i,j}_{pool},
	\label{eq:loss_local}
\end{equation}
where $y^{i,j}_{gb}$ represents the fine-grained domain label assigned by the PGBS module to the $j$-th local feature \( f^{i, j}_{pool} \), \( p^{i,{(u,v)}} \) represents the domain classifier’s prediction for the feature point \( f^{i,(u,v)} \) located at \( (u, v) \) within the feature \( f^i \), and \( f^{i, u, v} \) belongs to the feature points within the \( j \)-th local feature \( f^{i, j}_{pool} \).

\textbf{Global GB-DAL:} For the global features $f^{i}_{pool}$ of the image, we directly apply global average pooling on the feature map $f^{i}$. Similar to the local GB-DAL, the global GB-DAL also utilizes the PGBS module to obtain fine-grained domain labels  $y^{i}_{gb}$ for the entire image feature $f^{i}$. The optimization loss for the global GB-DAL can be expressed as:
\begin{equation}
	{\mathcal{L}}_{global}=-\sum_{i,u,v}y^{i}_{gb}\log{p^{i,{(u,v)}}},
	\label{eq:loss_global}
\end{equation}
where \( y^{i}_{gb} \) represents the fine-grained domain label assigned by the PGBS module to the entire feature map $f^{i}$, while \( p^{i,{(u,v)}} \) denotes the pixel-level predictions of the domain classifier for the feature map $f^{i}$.


\textbf{Total Loss:} After incorporating both local and global GB-DAL, the overall optimization loss of the object detection model can be defined as: 
\begin{equation}	\mathcal{L}=\mathcal{L}_{det}+\lambda\cdot(\mathcal{L}_{local}+\mathcal{L}_{global}),
	\label{eq:loss_total}
\end{equation}
where $\lambda$ is a trade-off parameter used to balance the detection task with the GB-DAL task, 
\( \mathcal{L}_{det} \) denotes the standard detection loss of Faster RCNN~\cite{ren2015faster}, comprising the detection head loss and the Region Proposal Network (RPN) loss, both of which consist of classification and regression losses.

\begin{figure}[tb]
	\centering
	\includegraphics[width=1\linewidth]{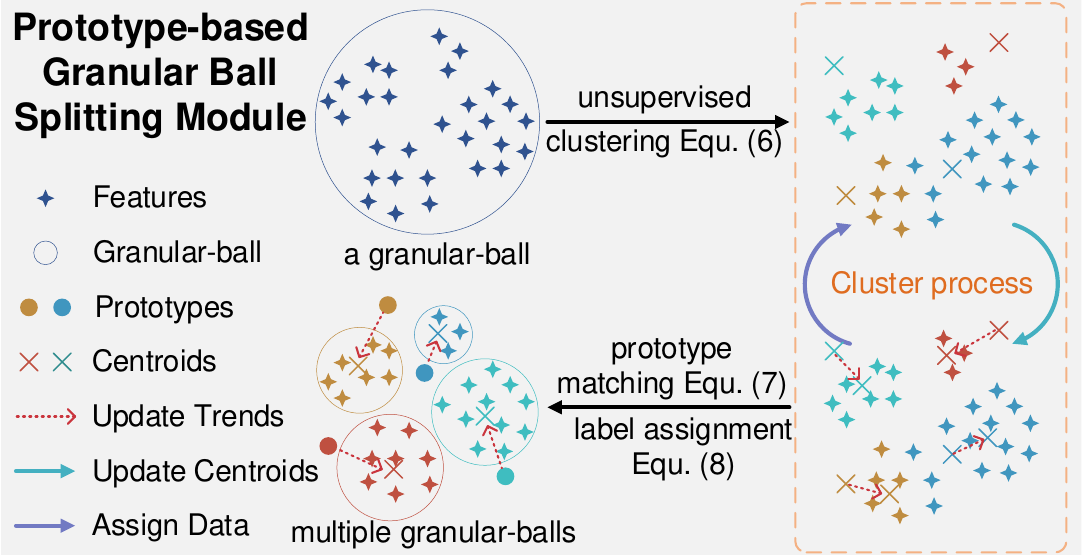}
	\caption{Illustration of the Prototype-based Granular Ball Splitting (PGBS) module, which conducts datasets split.
	}
	\label{fig:PGBS}
\end{figure}

\subsection{Prototype-based Granular Ball Splitting}
\label{sect33}
As shown in \figref{fig:PGBS}, the Prototype-based Granular Ball Splitting (PGBS) module uses unsupervised $K$-means clustering algorithm  for granular ball splitting \cite{xie2024gbg++,cheng2023fast,xia2019granular}.
Due to the random initialization of $K$ cluster centers, the resulting domain labels are arbitrary and unordered for each execution during training, which may cause training collapse. To ensure the consistency of domain labels throughout the entire training process, we employ a memory bank to store $K$ domain prototypes $q^k$ of the cluster centers. 

Specifically, we first employ $K$-means to cluster the local features $f^{i,j}_{pool}$, obtaining $K$ clusters $\{\mathcal{C}^1, \mathcal{C}^2,\cdots, \mathcal{C}^K\}$ and their centroids $\{c^1, c^2, \cdots, c^K\}$. The overall objective of the clustering process is to minimize the distance of each sample $f^{i,j}_{pool}$ within a cluster $\mathcal{C}^{k}$ to the cluster center $c^{k}$, \ie, to minimize the following loss function:
\begin{equation}
J=\sum_{k=1}^{K}\sum_{f^{i,j}_{pool}\in \mathcal{C}^{k}}\|f^{i,j}_{pool}-c^{k}\|_{2},
	\label{eq:loss_cluster}
\end{equation}
where \( \|\cdot\|_2 \) denotes the \( \ell_2 \)-norm. We utilize the \( \ell_2 \) distance to measure the distance between samples and cluster centers.
Subsequently, to determine the optimal bipartite matching between the cluster centers $c$ and domain prototypes $q$, we search for the permutation $\sigma$ of $K$ elements with the minimum cost:
\begin{equation}
	\hat{\sigma}=\operatorname*{argmin}_{\sigma\in\Theta_K}\sum_{k=1}^K{D}_{match}(q^k,c^{\sigma(k)}),
	\label{eq:match}
\end{equation}
where ${D}_{match}(q^k,c^{\sigma(k)})$ is a pairwise matching \( \ell_2 \) distance between domain prototype $q^k$ and a cluster center with index $\sigma(k)$, and $\Theta_K$ is the permutations of $K$ elements. The optimal assignment $\hat{\sigma}$ can be efficiently obtained through the Hungarian algorithm.
With the assistance of $\hat{\sigma}$, we can assign domain label $y^{i,j}_{gb}$ to each feature $f^{i,j}_{pool}$ of the image via the formula:
\begin{equation}
	y^{i,j}_{gb}=k \quad \text{if } {f^{i,j}_{pool}\in \mathcal{C}^{\hat{\sigma}(k)}},
	\label{eq:assign_label}
\end{equation}

Regarding the initialization and updating of domain prototypes, at the beginning of training, we directly utilize the cluster center features to initialize the domain prototypes. In the subsequent iterations, we update the domain prototypes $q^k$ through exponential moving average (EMA) in combination with cluster center features $c^{\hat{\sigma}(k)}$. The domain prototypes at iteration $t$ can be expressed as:
\begin{equation}
	q_{t}^{k}=\alpha\cdot q_{t-1}^{k}+(1-\alpha)\cdot c_{t}^{\hat{\sigma}(k)},
	\label{eq:ema}
\end{equation}
where \( q_{t}^{k} \) represents the domain prototype at the current \( t \)-th iteration, while \( q_{t-1}^{k} \) denotes the domain prototype from the previous iteration. The parameter $\alpha$ serves as a regulating factor that controls the update rate of the prototype, where $\alpha$ is set as 0.99. At \( t = 0 \), the initial prototype is set as \( q_{0}^{k} = c_{0}^{k} \).

Additionally, to balance the weight of clustering center features in the domain prototypes across iterations, we linearly increase the regulating factor $\alpha$ within the range [0.5, 0.99] during the first 100 iterations.

\begin{algorithm}[t]
    \caption{Optimization process of the model.}
    \label{alg:optimization_process}
    \begin{algorithmic}[1] 
        \REQUIRE training samples $D_{train}=\{{x}_{j}, y_j\}_{j=1}^{N}$ from multiple source datasets. 
        \ENSURE detection model with backbone $E$ (feature encoder) and detection head $H$.
        \FOR{a batch $\mathcal{X}=\{{x}_{b}\}_{b=1}^{B}$ in $D_{train}$}
        \STATE \textbf{SNF:} \textcolor{cyan}{\textit{\# non-causal factors simulation phase}}
        \STATE Forward computation to obtain $\mathcal{L}_{cls}$ via \equref{eq:adversarial_cls}.
        \STATE Perform backpropagation based on $\mathcal{L}_{cls}$ to compute $adv$ using \equref{eq:adversarial_pert} and generate $x_{adv}$ via \equref{eq:adversarial}.
        \STATE \textbf{GB-DAL:} \textcolor{cyan}{\textit{\# causality invariant learning phase}}
        \STATE Feed both $x$ and $x_{adv}$ into $E$ to extract features $f$.
        \STATE Use the PGBS module to obtain the dense domain label $y_{gb}$ for feature $f$ via \equref{eq:loss_cluster}, \equref{eq:match} and \equref{eq:assign_label}.
        \STATE Compute ${\mathcal{L}}_{local}$  and ${\mathcal{L}}_{global}$  via Eq. (\ref{eq:loss_local}) and (\ref{eq:loss_global}) for fine-grained domain alignment. 
        \STATE \textbf{Supervised Training:} \textcolor{cyan}{\textit{\# detection head training}}
        \STATE Feed the features $f$ into $H$ to compute the standard detection loss $\mathcal{L}_{det}$.
        \STATE Update model via total loss in \equref{eq:loss_total}.
        
        
        \ENDFOR
    \end{algorithmic}
\end{algorithm}

\subsection{Simulated Non-causal Factors}
\subsubsection{The Proposed SNF}
%
Given that both adversarial perturbations (attacks) and non-causal factors share two similar intuitive characteristics: 1) there is no causal association with the category, but degrades the model performance and 2) human imperceptible, 
We propose to simulate non-causal factors using adversarial perturbations as a means of data augmentation, thereby reducing their implicitness and facilitating GB-DAL to effectively mine and filter out implicit non-causal factors.

In detail, we adhere to a common adversarial training strategy, such as the Fast Gradient Sign Method (FGSM~\cite{goodfellow2014explaining}), to obtain adversarial samples injected with non-causal factors (\ie, adversarial perturbations). 
First, the clean training sample \( x_{cle} \) is fed into the model for forward propagation to compute the classification loss: 
\begin{equation}
	 \mathcal{L}_{cls} = -\sum_i y_i \log p_i , 
	\label{eq:adversarial_cls}
\end{equation} 
where \( p_i \) represents the predicted class probability from the classification branch of the ROI Head, and \( y_i \) denotes the class label. Subsequently, the gradient-based adversarial perturbation (\ie, adv) is obtained through backpropagation: 
\begin{equation}
	 adv= -\epsilon\cdot{sign}(\nabla_{x_{cle}}\mathcal{L}_{cls}), 
	\label{eq:adversarial_pert}
\end{equation}
where \(\nabla_{x_{cle}}\mathcal{L}_{cls}\) represents the backpropagated gradient of the classification loss \(\mathcal{L}_{cls}\) with respect to the clean sample \(x_{cle}\), $\epsilon$ is a small perturbation coefficient, and $sign(\cdot)$ means the Sign function.

Although the perturbation generated by the above process has no causal relationship with the class label, it establishes an association with the class label through the model, since injecting this perturbation into the image significantly increases the model's confidence to predict a designated class label except \( y_i \). To further enhance the diversity of the simulated non-causal factors, the class label  \( y_i \) in \equref{eq:adversarial_cls} is replaced with random labels instead of the true labels.

By adding perturbations to clean samples, we get adversarial samples: 
\begin{equation}
	x_{adv}=x_{cle}+adv. 
	\label{eq:adversarial}
\end{equation} 
Finally, the clean and adversarial samples are jointly fed as augmented datasets into the object detection model with GB-DAL, to facilitate causal feature learning. 

The complete algorithm for domain generalization object detection (DGOD) via GB-DAL and SNF is summarized in \algref{alg:optimization_process}.

\subsubsection{Plausibility Analysis of SNF}

\begin{table}[t]
\centering
\caption{Performance quantification of SNF on NICO \cite{he2021towards}.}
\label{tab:nicocomp}
\begin{tabular}{l|ccc}
\hline
        Model & Resnet18~\cite{he2016deep} & EiHi Net \cite{wei2022eihi} & FACT~\cite{xu2021fourier} \\
        \hline
        \textit{without} SNF & 28.6 & 37.4 & 35.2 \\
        \hline
        \textit{with} SNF & \textbf{30.2} & \textbf{39.5} & \textbf{36.4} \\
        \hline
\end{tabular}
\end{table}

\begin{table*}[h]
\begin{center}
\renewcommand{\arraystretch}{0.9}
\setlength{\tabcolsep}{3mm}{
\begin{tabular}{l|c c c c c c c c c}
\hline
\textbf{Method} & bike & bus & car & motor & person & rider & train & truck & mAP \\
\hline

\multicolumn{10}{c}{\textbf{C\&B$\to$F}} \\
\hline
ERM & 35.4 & 36.9 & 46.2 & 26.8 & 31.9 & 45.0 & 4.0 & 25.8 & 31.5 \\
DAL \cite{ganin2015unsupervised} & 33.6 & 33.1 & 35.4 & 23.2 & 26.3 & 39.7 & 9.1 & 15.6 & 27.0 \\
\textcolor{black}{DivAlign w/o Div}& \textcolor{black}{31.9} & \textcolor{black}{37.8} & \textcolor{black}{44.3} & \textcolor{black}{28.7} & \textcolor{black}{27.5} & \textcolor{black}{40.2} & \textcolor{black}{20.8} & \textcolor{black}{26.6} & \textcolor{black}{32.2} \\
\textbf{GB-DAL}(ours) & 35.6 & 43.1 & 47.8 & 31.2 & 32.8 & 43.5 & 12.4 & 31.8 & \textbf{34.8} \\ 
\cline{1-10}
FACT~\cite{xu2021fourier} & 30.3 & 33.4 & 41.3 & 24.7 & 29.0 & 42.1 & 2.4 & 20.6 & 28.0 \\
FACT+\textbf{SNF} & 33.4 & 35.4 & 42.2 & 24.9 & 29.5 & 41.2 & 4.9 & 22.1 & 29.2(\textbf{+1.2}) \\
FSDR~\cite{huang2021fsdr} & 35.0 & 38.7 & 45.9 & 27.4 & 32.3 & 45.1 & 13.9 & 28.1 & 33.3 \\
FSDR+\textbf{SNF} & 37.9 & 41.1 & 47.3 & 29.7 & 33.0 & 47.8 & 18.8 & 25.1 & 35.1(\textbf{+1.8}) \\
NP \cite{fan2023towards} & 38.2 & 42.9 & 50.2 & 29.9 & 33.2 & 47.5 & 16.7 & 25.5 & 35.5 \\
NP+\textbf{SNF} & 38.4 & 47.3 & 50.8 & 32.1 & 34.6 & 47.7 & 31.1 & 27.6 & 38.7(\textbf{+3.2}) \\
OA-DG \cite{lee2024object} & 38.8 & 40.4 & 53.8 & 33.8 & 34.9 & 47.1 & 21.3 & 31.4 & 37.7 \\
OA-DG+\textbf{SNF} & 38.9 & 41.6 & 54.2 & 33.7 & 35.2 & 47.8 & 22.4 & 33.8 & 38.5(\textbf{+0.8}) \\
DivAlign \cite{danish2024improving} & 39.2 & 42.4 & 52.0 & 32.0 & 33.1 & 48.3 & 27.6 & 34.4 & 38.6 \\
DivAlign+\textbf{SNF} & 39.5 & 43.5 & 51.5 & 33.8 & 33.9 & 48.6 & 27.4 & 35.2 & 39.2(\textbf{+0.6}) \\
 \hline
\textbf{GB-DAL+SNF}(ours) & 38.5 & 44.6 & 51.8 & 34.6 & 34.2 & 47.9 & 24.7 & 36.5 & 39.1(\textbf{+4.3}) \\
\textcolor{black}{\textbf{GB-DAL+Div+SNF}(ours)} & \textcolor{black}{38.8} & \textcolor{black}{44.8} & \textcolor{black}{53.3} & \textcolor{black}{34.8} & \textcolor{black}{34.5} & \textcolor{black}{48.6} & \textcolor{black}{25.4} & \textcolor{black}{36.2} & \textcolor{black}{\textbf{39.6}(\textbf{+0.5})} \\
\cline{1-10}
Oracle - Train on target  & 37.8 & 47.4 & 53.0 & 31.6 & 52.9 & 34.3 & 37.0 & 40.6 & 41.8 \\
\hline

\multicolumn{10}{c}{\textbf{F\&B$\to$C}} \\
\hline
ERM & 37.1 & 45.2 & 53.6 & 32.7 & 37.7 & 47.8 & 5.4 & 38.2 & 37.2 \\
DAL \cite{ganin2015unsupervised} & 36.2 & 44.0 & 51.9 & 30.6 & 34.8 & 44.1 & 19.0 & 23.9 & 35.6 \\
\textcolor{black}{DivAlign w/o Div} & \textcolor{black}{35.6} & \textcolor{black}{47.4} & \textcolor{black}{48.5} & \textcolor{black}{32.8} & \textcolor{black}{35.2} & \textcolor{black}{41.5} & \textcolor{black}{25.4} & \textcolor{black}{37.1} & \textcolor{black}{38.0} \\
\textbf{GB-DAL}(ours) & 39.2 & 50.1 & 53.8 & 36.3 & 39.8 & 48.0 & 30.7 & 40.3 & \textbf{42.3} \\
\cline{1-10}
FACT~\cite{xu2021fourier} & 36.2 & 41.7 & 53.3 & 30.0 & 34.5 & 45.7 & 2.8 & 34.3 & 34.8 \\
FACT+\textbf{SNF} & 38.0 & 54.4 & 56.9 & 36.1 & 38.4 & 46.8 & 6.9 & 37.5 & 39.4(\textbf{+4.6}) \\
FSDR~\cite{huang2021fsdr} & 36.0 & 47.0 & 54.9 & 34.9 & 37.1 & 46.8 & 5.0 & 33.1 & 36.9 \\
FSDR+\textbf{SNF} & 39.0 & 55.5 & 55.0 & 38.8 & 38.7 & 48.4 & 12.2 & 38.7 & 40.8(\textbf{+3.9}) \\
NP \cite{fan2023towards} & 37.5 & 44.9 & 53.3 & 30.2 & 35.0 & 47.0 & 5.0 & 33.6 & 35.8 \\
NP+\textbf{SNF} & 41.4 & 55.5 & 55.7 & 39.9 & 38.1 & 47.4 & 14.9 & 38.0 & 41.4(\textbf{+5.6}) \\
OA-DG \cite{lee2024object} & 39.5 & 53.1 & 55.9 & 40.4 & 40.6 & 51.0 & 25.6 & 41.2 & 43.4 \\
OA-DG+\textbf{SNF} & 40.8 & 54.2 & 56.6 & 40.9 & 40.9 & 51.8 & 26.8 & 41.5 & 44.2(\textbf{+0.8}) \\
DivAlign \cite{danish2024improving} & 41.8 & 54.7 & 59.1 & 35.2 & 41.2 & 47.5 & 32.4 & 42.7 & 44.3 \\
DivAlign+\textbf{SNF} & 41.8 & 54.9 & 58.8 & 37.2 & 41.3 & 48.3 & 32.3 & 42.8 & 44.7(\textbf{+0.4}) \\
 \hline
\textbf{GB-DAL+SNF}(ours) & 43.7 & 55.1 & 57.8 & 38.5 & 41.3 & 51.1 & 30.8 & 40.8 & 44.9(\textbf{+2.6}) \\

\textcolor{black}{\textbf{GB-DAL+Div+SNF}(ours)} & \textcolor{black}{44.1} & \textcolor{black}{55.3} & \textcolor{black}{58.2} & \textcolor{black}{38.8} & \textcolor{black}{41.2} & \textcolor{black}{51.0} & \textcolor{black}{31.5} & \textcolor{black}{41.5} & \textcolor{black}{\textbf{45.2}(\textbf{+0.3})} \\
\cline{1-10}
Oracle - Train on target  & 43.7 & 59.9 & 58.6 & 39.1 & 39.8 & 53.1 & 34.5 & 36.9 & 45.7 \\
\hline

\multicolumn{10}{c}{\textbf{C\&F$\to$B}} \\
\hline
ERM & 25.2 & 18.4 & 41.5 & 12.9 & 29.7 & 27.6 & - & 18.6 & 24.8 \\
DAL \cite{ganin2015unsupervised} & 26.6 & 18.3 & 41.5 & 13.0 & 30.2 & 27.9 & - & 18.1 & 25.1 \\
\textcolor{black}{DivAlign w/o Div} & \textcolor{black}{25.1} & \textcolor{black}{17.2} & \textcolor{black}{40.8} & \textcolor{black}{15.1} & \textcolor{black}{30.3} & \textcolor{black}{28.6} & \textcolor{black}{-} & \textcolor{black}{19.5} & \textcolor{black}{25.2} \\
\textbf{GB-DAL}(ours) & 27.9 & 19.6 & 40.2 & 14.4 & 34.6 & 30.1 & - & 18.0 & \textbf{26.4 }\\
\cline{1-10}
FACT~\cite{xu2021fourier} & 28.0 & 18.6 & 43.9 & 11.7 & 31.5 & 29.1 & - & 20.6 & 26.2 \\
FACT+\textbf{SNF} & 28.2 & 17.1 & 44.6 & 12.8 & 32.0 & 30.8 & - & 19.3 & 26.4(\textbf{+0.2}) \\
FSDR~\cite{huang2021fsdr} & 28.1 & 18.7 & 45.9 & 13.3 & 32.1 & 29.6 & - & 19.6 & 26.8 \\
FSDR+\textbf{SNF} & 28.9 & 19.9 & 45.0 & 14.3 & 32.4 & 29.7 & - & 20.6 & 27.3(\textbf{+0.5}) \\
NP \cite{fan2023towards} & 29.8 & 17.8 & 44.0 & 13.3 & 32.6 & 28.9 & - & 20.7 & 26.7 \\
NP+\textbf{SNF} & 29.8 & 21.2 & 44.8 & 15.6 & 34.4 & 32.1 & - & 22.4 & 28.6(\textbf{+1.9}) \\
OA-DG \cite{lee2024object} & 30.5 & 20.5 & 41.9 & 15.6 & 33.7 & 29.4 & - & 20.1 & 27.4 \\
OA-DG +\textbf{SNF} & 30.8 & 20.9 & 41.5 & 16.2 & 34.3 & 30.8 & - & 20.5 & 27.9(\textbf{+0.5}) \\
DivAlign \cite{danish2024improving} & 29.7 & 18.7 & 42.4 & 16.9 & 35.7 & 31.9 & - & 21.3 & 28.1 \\

DivAlign+\textbf{SNF} & 29.8 & 19.3 & 43.1 & 16.6 & 35.9 & 32.3 & - & 21.5 & 28.4(\textbf{+0.3}) \\
 \hline
\textbf{GB-DAL+SNF}(ours) & 31.2 & 20.4 & 44.4 & 13.3 & 34.1 & 29.7 & - & 20.9 & 27.7(\textbf{+1.3}) \\

\textcolor{black}{\textbf{GB-DAL+Div+SNF}(ours)} & \textcolor{black}{31.8} & \textcolor{black}{21.1} & \textcolor{black}{44.8} & \textcolor{black}{15.8} & \textcolor{black}{34.7} & \textcolor{black}{31.6} & \textcolor{black}{-} & \textcolor{black}{21.5} & \textcolor{black}{\textbf{28.8}(\textbf{+1.1})} \\
\cline{1-10}
Oracle - Train on target  & 37.0 & 42.6 & 59.3 & 27.8 & 39.5 & 40.3 & - & 39.2 & 40.8 \\
\hline

\end{tabular}}
\end{center}

\caption{Results (\%) under multi-source DGOD setting.
		Models are trained on two datasets from \{\textbf{C}, \textbf{F}, \textbf{B}\} and tested on the rest, alternatively. ERM,  DAL, DivAlign w/o div, and GB-DAL denote augmentation-free methods, without using any data augmentation methods. The remaining methods use different data augmentation approaches.
        }
\label{tab:main}
\end{table*}

This subsection presents the plausibility analysis about the intuition of SNF by initiating a question: \textit{whether SNF is detrimental to causal factors}? 

\begin{figure}[t]
	\centering
	\includegraphics[width=1\linewidth]{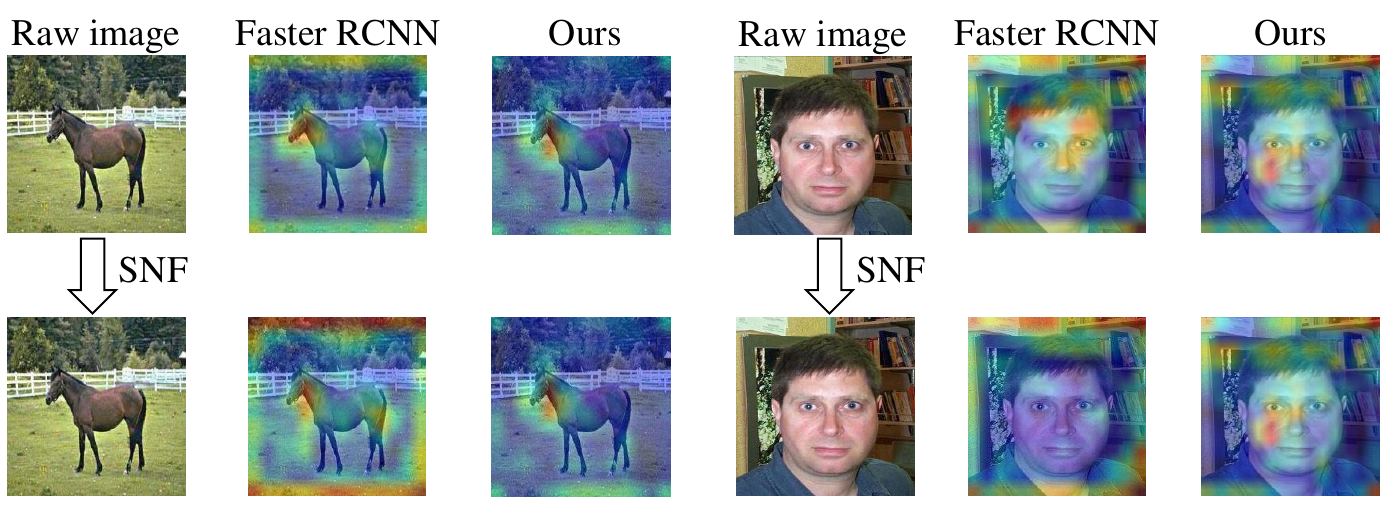}
	\caption{Comparisons on raw clean samples (the first row) and the adversarial samples (the second row). 
	}
	\label{fig:sc}
\end{figure}

\textbf{SNF helps improve causal representation}. 
From the perspective of human recognition mechanisms, since the adversarial perturbations introduced by SNF do not affect normal human recognition, SNF does not bring negative impact to causal factors. Additionally, we conduct an experiment by evaluating both clean and adversarial samples using Faster RCNN \cite{ren2015faster} and our proposed model. As can be seen in the first row of \figref{fig:sc}, we observe that SNF helps improve causal representation and the proposed model can better focus on causal parts compared to Faster RCNN. 

To further validate this claim, we conduct a quantitative classification experiment on the NICO benchmark\cite{he2021towards}, a particular dataset with spurious correlations for out-of-distribution (OOD) image classification, which contains images of animals and vehicles in various contexts, with each category represented by 10 different contexts. In this experiment, the NICO-animal is used. We train the model on 8 contexts and test it on the remaining 2 contexts.  Tab.  \ref{tab:nicocomp} demonstrates that the SNF module effectively filters out non-causal factors, thereby enhancing causality.


\textbf{SNF brings adversarial robustness as a free-lunch}. By revisiting \figref{fig:sc}, we find that the proposed SNF enables the GB-DAL based object detector to achieve adversarial robustness, as a free-lunch. The results in the second row show that our model is clearly better than Faster RCNN when tested on the adversarial samples. 
From the perspective of adversarial training~\cite{goodfellow2014explaining}, the proposed GB-DAL plugged with SNF shown in \figref{fig:overview} can be viewed as a special adversarial training framework of Faster RCNN object detector.
In other words, the proposed SNF unexpectedly achieves both domain generalization and adversarial robustness, as corroborated by the quantitative experiments in \tabref{tab:adv_gauss}. It is noteworthy that SNF also exhibits robustness to natural noise, indicating the robustness of our approach for open-world scenarios.

\section{Experiments}
\label{sec:experiments}

\begin{table*}[tb]
	\begin{center}
		\setlength{\tabcolsep}{1.4mm}{
			\begin{tabular}{p{0.4mm} l|c|c c c c c c c c|c }
				\hline
				\multicolumn{2}{ l|}{Methods}                                       & Training data      & bike        & bus         & car           & motor         & person           & rider         & train         & truck          & mAP \\
				\hline
				\hline
				\multicolumn{1}{ l|}{\multirow{8}*{DA}}& DAF \cite{chen2018domain}   &                   & 32.6          & 41.3          & 42.8          & 28.9          & 31.6          & 43.6          & 21.2          & 23.6          & 33.2 \\
				\multicolumn{1}{ l|}{}& SW-DA \cite{saito2019strong}                 &                   & 35.8          & 43.8          & 48.9          & 28.9          & 31.8          & 44.3          & 28.0          & 21.0          & 35.3 \\
				\multicolumn{1}{ l|}{}& SC-DA \cite{zhu2019adapting}                 & Source (\textbf{C})     & 34.5          & 42.5          & 52.1          & 29.2          & 33.8          & 42.1          & 26.5          & 26.8          & 35.9 \\
				\multicolumn{1}{ l|}{}& MTOR \cite{cai2019exploring}                 & \&                & 35.6          & 38.6          & 44.0          & 28.3          & 30.6          & 41.4          & 40.6          & 21.9          & 35.1 \\
				\multicolumn{1}{ c|}{}& ICR-CCR \cite{xu2020exploring}               & Target (\textbf{F})     & 34.6          & 45.1          & 49.2          & 30.3          & 32.9          & 43.8          & 36.4          & 27.2          & 37.4 \\
				\multicolumn{1}{ l|}{}& Coarse-to-Fine \cite{zheng2020cross}         & (without labels)  & 37.4          & 43.2 & 52.1          & \textbf{34.7} & 34.0          & \textbf{46.9}          & 29.9          & \textbf{30.8}          & 38.6 \\
				\multicolumn{1}{ l|}{}& GPA \cite{xu2020cross}                       &                   & \textbf{38.7}          & \textbf{45.7}          & 54.1          & 32.4          & 32.9 & 46.7 & \textbf{41.1} & 24.7 & 39.5 \\
				\multicolumn{1}{ l|}{}& Center-Aware \cite{hsu2020every}             &                   & 36.1 & 44.9          & \textbf{57.1} & 29.0          & \textbf{41.5}          & 43.6          & 39.7          & 29.4          & \textbf{40.2} \\
				\hline
				\hline
				\multicolumn{1}{ l|}{\multirow{10}*{DG}}& FACT \cite{xu2021fourier}   & \multirow{10}*{Multiple Sources}    & 30.3          & 33.4          & 41.3          & 24.7          & 29.0          & 42.1          & 2.4           & 20.6          & 28.0 \\
				\multicolumn{1}{ l|}{}& ERM                                         &                                   & 35.4          & 36.9          & 46.2          & 26.8          & 31.9          & 45.0          & 4.0           & 25.8          & 31.5 \\
				\multicolumn{1}{ l|}{}& FSDR \cite{huang2021fsdr}                    &                                   & 35.0          & 38.7          & 45.9          & 27.4          & 32.3          & 45.1          & 13.9          & 28.1          & 33.3 \\
				\multicolumn{1}{ l|}{}& DIDN \cite{lin2021domain}                    &                                   & 31.8          & 38.4          & 49.3          & 27.7          & \textbf{35.7} & 26.5          & {24.8} & 33.1          & 33.4 \\
				\multicolumn{1}{ l|}{}& \textbf{GB-DAL}(ours)                         &                                   & 35.6          & 43.1          & 47.8          & 31.2          & 32.8          & 43.5          & 12.4          & 31.8          & 34.8 \\
				\multicolumn{1}{ l|}{}& NP \cite{fan2023towards}               &                                   & 38.2          & 42.9          & 50.2          & 29.9          & 33.2          & 47.5          & 16.7          & 25.5          & 35.5 \\

\multicolumn{1}{ l|}{} & OA-DG \cite{lee2024object} &  & 38.8 & 40.4 & 53.8 & 33.8 & 34.9 & 47.1 & 21.3 & 31.4 & 37.7 \\ 
\multicolumn{1}{ l|}{} & DivAlign \cite{danish2024improving} &  & 39.2 & 42.4 & 52.0 & 32.0 & 33.1 & {48.3} & \textbf{27.6} & 34.4 & 38.6 \\ 

				\multicolumn{1}{ l|}{}& \textbf{GB-DAL+SNF}(ours)                               &                                   & {38.5} & {44.6} & {51.8} & {34.6} & 34.2          & {47.9} & 24.7          & \textbf{36.5} & {39.1} \\
				
				\multicolumn{1}{ l|}{}&\textcolor{black}{\textbf{GB-DAL+Div+SNF}(ours)}& & \textcolor{black}{\textbf{38.8}} & \textcolor{black}{\textbf{44.8}} & \textcolor{black}{\textbf{53.3}} & \textcolor{black}{\textbf{34.8}} & \textcolor{black}{34.5} & \textcolor{black}{\textbf{48.6}} & \textcolor{black}{25.4} & \textcolor{black}{36.2} & \textcolor{black}{\textbf{39.6}} \\
                \hline
                \multicolumn{2}{ l|}{Oracle - Train on target}                      & Target            & 37.8          & 47.4          & 53.0          & 31.6          & 52.9          & 34.3          & 37.0          & 40.6          & 41.8 \\
                
				\hline
		\end{tabular}}
	\end{center}
	\caption{Results tested on Foggy Cityscapes (\textbf{F}).
		DAOD methods are trained on Cityscapes (\textbf{C}) and unlabeled Foggy Cityscapes (\textbf{F}).
		DGOD methods are trained on \textbf{C} and \textbf{B}.
		The best AP in each class and mAP are highlighted in bold.
	}
	\label{tab:comparision}
\end{table*}

\subsection{Dataset}
We utilize six object detection benchmarks, including Cityscapes \cite{cordts2016cityscapes}, Foggy Cityscapes \cite{sakaridis2018semantic}, Rain Cityscapes \cite{Hu_2019_CVPR}, SIM 10k \cite{johnson2016driving}, PASCAL VOC \cite{everingham2015pascal}, and BDD100k \cite{yu2020bdd100k}, to construct cross-domain experimental scenarios for evaluating the model's domain generalization capability in unseen domains.

\textbf{Cityscapes} \cite{cordts2016cityscapes} is a large-scale dataset focused on urban scenes, collected from street views in 50 different cities, primarily captured during daytime and under favorable weather conditions.

\textbf{Foggy Cityscapes} \cite{sakaridis2018semantic} is a foggy image dataset synthesized based on the Cityscapes dataset. The synthesis process employs a physical model to simulate foggy conditions for each image using depth maps. The dataset maintains the same number of images, annotation files, and partitioning scheme for the training, validation, and test sets as the Cityscapes dataset. 
Both the Cityscapes dataset and the Foggy Cityscapes dataset contain 2,975 images for training and 500 images for testing. 

\textbf{Rain Cityscapes} \cite{Hu_2019_CVPR} is a synthetic dataset based on Cityscapes, simulating various rainy conditions by adding rain streaks and haze. It contains 9,432 training images and 1,188 test images, with annotations inherited from Cityscapes. 

\textbf{SIM10K} \cite{johnson2016driving}  is a synthetic object detection dataset generated using a game engine, consisting of 10,000 images, with 5,000 images allocated for the training set and the remaining 5,000 for the test set. 

\textbf{PASCAL VOC}~\cite{everingham2015pascal} is collected from the real world, in which 5,011 pictures are used for training and 4,952 pictures are used for testing.

\textbf{BDD100k}~\cite{yu2020bdd100k} is a large-scale dataset for autonomous driving, encompassing diverse weather, lighting, and road conditions. We use 10,000 images as the train set and 10,000 images as the test set. 

For convenience, we abbreviate \underline{S}IM 10k, \underline{C}ityscapes, \underline{F}oggy Cityscapes, \underline{B}DD100k, \underline{R}ain Cityscapes, \underline{P}ASCAL VOC as \textbf{S, C, F, B, R, P}, respectively. 

\subsection{Experimental Setup}
\textbf{Implementation Details.}
To verify the effectiveness of our GB-DAL and SNF methods,
we conduct both single and multi-source DG experiments.
For each source and target domain pair, we only evaluate the result on the common classes shared in their label spaces. 
To unify the annotation styles of these datasets, we regard labels \{\emph{motor}, \emph{motorcycle} and \emph{motorbike}\} as \{\emph{motor}\} and labels \{\emph{bike} and \emph{bicycle}\} as \{\emph{bike}\}.
Before training, we resize the image to the short side of 600 pixels while keeping the aspect ratio of the images unchanged.
We train the model for a total of 10 epochs. During the training process, the initial learning rate is set to 0.002, and starting from the 7th epoch, the learning rate is decayed to 0.0002 to facilitate better convergence of the model.
The batch size is set to 1 for all datasets. In our experiments, training is conducted using a single NVIDIA TITAN XP GPU. The mean average precision (mAP) is reported with an Intersection over Union (IoU) threshold of 0.5. 

\textbf{Baseline.}
We build our method upon Faster RCNN~\cite{ren2015faster}, using VGG-16 pre-trained on ImageNet~\cite{russakovsky2015imagenet} as the backbone, and employ Stochastic Gradient Descent (SGD)~\cite{robbins1951stochastic} as the optimizer.


\subsection{Experimental Results}
To validate the effectiveness of our proposed GB-DAL and SNF in enhancing model domain generalization, we conduct multi-source and single-source domain generalization object detection (DGOD) experiments. Additionally, we perform a multi-source domain generalization experiment on image classification tasks.

\textbf{Multi-source DGOD}. 
For our multi-source domain generalization object detection experiments, we categorize the baseline methods into two groups to ensure fair comparison, including augmentation-free methods and augmentation-inclusive methods. The first group comprises methods without any data augmentation, including standard ERM, DAL~\cite{ganin2015unsupervised}, and our proposed GB-DAL. To establish a fair baseline for evaluating a augmentation-free DivAlign framework, we specifically introduce \textbf{DivAlign w/o Div} as a comparative baseline, which represents a variant of the SOTA DivAlign~\cite{danish2024improving} with its diversification augmentation component removed, enabling direct comparison of alignment mechanisms under identical augmentation-free conditions. The second group consists of methods with data augmentation, encompassing current mainstream DGOD approaches: FACT~\cite{xu2021fourier}, FSDR~\cite{huang2021fsdr}, NP~\cite{fan2023towards}, OA-DG~\cite{lee2024object}, and DivAlign~\cite{danish2024improving} . Furthermore, we test our proposed SNF module with all methods to assess its plug-and-play effectiveness and synergistic potential. 
\tabref{tab:main} presents the results of multi-source domain generalizable object detection, which includes three multi-source domain generalization experiments: \textbf{C}\&\textbf{B}$\to$\textbf{F}, \textbf{F}\&\textbf{B}$\to$\textbf{C}, and \textbf{C}\&\textbf{F}$\to$\textbf{B}. 
In the augmentation-free methods, our proposed GB-DAL demonstrates superior performance compared to conventional DAL and other non-augmented baselines. Specifically, GB-DAL achieves mAP scores of 34.8\%, 42.3\% and 26.4\% across the three tasks (C\&B$\to$F, F\&B$\to$C and C\&F$\to$B), representing improvements of 7.8\%, 6.7\% and 1.3\% over conventional DAL, respectively. More importantly, GB-DAL consistently outperforms {\textit{DivAlign w/o Div}} (32.2\%, 38.0\% and 25.2\%), which serves as a fair baseline for comparison under identical augmentation-free conditions. These results provide direct evidence that, when evaluated on equal condition without augmentation, GB-DAL's granular alignment mechanism is more effective than the alignment component of the SOTA method. In the augmentation-inclusive methods, \textit{GB-DAL+Div+SNF}, which combines the proposed GB-DAL with DivAlign's augmentation and the proposed SNF module, achieves the best overall performance on all tasks (C\&B$\to$F: 39.6\%, F\&B$\to$C: 45.2\% and C\&F$\to$B: 28.8\%). This result provides compelling evidence that our augmentation-free alignment framework, GB-DAL, is not only effective as a standalone solution but also offers complementary value that can be integrated with existing augmentation strategies to achieve further enhancements in performance. 


Additionally, the proposed SNF module demonstrates strong plug-and-play capabilities across multiple state-of-the-art DG methods. As shown in \tabref{tab:main}, SNF consistently improves performance in all three cross-domain tasks. In C\&B→F, it enhances FACT, FSDR, NP, OA-DG, and DivAlign by +1.2\%, +1.8\%, +3.2\%, +0.8\%, and +0.6\% in mAP, respectively. On the more challenging F\&B→C task, the gains are even more pronounced: +4.6\% (FACT), +3.9\% (FSDR), +5.6\% (NP), while it still improves OA‑DG and DivAlign by +0.8\% and +0.4\%. For C\&F→B, modest but consistent improvements are observed across all methods (e.g., +1.9\% for NP and +0.5\% for OA‑DG). These results confirm that SNF effectively augments a wide variety of DG approaches, validating its versatility and general‑purpose utility as a plug‑and‑play module.
{

}

\tabref{tab:comparision} shows the experimental results of the comparison between our proposed method and domain adaptation (DA) and domain generalization (DG) methods. It is noteworthy that DA methods can access to unlabeled target domain data during the training phase, whereas DG methods are entirely unaware of any target domain information. The experimental results demonstrate that our method achieves the bset performance among DG approaches and even surpasses some DA methods, thereby substantiating the superiority of our proposed approach. Although a performance gap persists when compared to certain DA methods such as GPA \cite{xu2020cross} and Center-Aware \cite{hsu2020every}, this discrepancy is justifiable as these DA methods benefit from the utilization of target domain data during their training process.

\begin{table}[t]
	\begin{center}
		\setlength{\tabcolsep}{1.3mm}{
			\begin{tabular}{l|c c c c c c}
				\hline
				Method                                    & \textbf{F}          & \textbf{R}          & \textbf{B}    & \textbf{S}  & \textbf{P}   \\
				\hline
				ERM                                & 35.9                & 49.6                & 41.3          & 39.2          & 59.1 \\
				FACT~\cite{xu2021fourier}                 & 35.8                & 48.8                & 42.0          & 41.2          & 63.0 \\
				FSDR~\cite{huang2021fsdr}                 & 43.3                & 52.7                & 45.0          & 42.2          & 58.7 \\
				FACT+\textbf{SNF} (ours)                         & 36.1                & 53.0                & \textbf{45.3} & 42.0          & 63.9 \\
				FSDR+\textbf{SNF} (ours)                        & 43.7                & 53.3                & 45.2          & \textbf{43.9} & 63.1 \\
				\textbf{GB-DAL+SNF (ours)}                  & \textbf{48.5}       & \textbf{53.9}       & 42.4          & 43.2          & \textbf{65.5} \\
				\hline
		\end{tabular}}
	\end{center}
	\caption{Results of the single-source DGOD setting, which are trained on \textbf{C} and tested on \textbf{F}, \textbf{R}, \textbf{B}, \textbf{S} and \textbf{P}, respectively for the shared category of \{\emph{car}\}.
	}
	\label{tab:single}
\end{table}

\textbf{Single-source DGOD}. To further validate the effectiveness of the proposed method in improving model domain generalization performance, we conducted a single-source DGOD experiment. \tabref{tab:single} presents the experimental results where the model was trained on dataset \textbf{C} and tested on target domain datasets \textbf{F}, \textbf{R}, \textbf{B}, \textbf{S}, and \textbf{P}. The results demonstrate that our method achieves the best generalization performance across multiple target domains, with mAP scores of 48.5\%, 53.9\%, and 65.5\% on \textbf{F}, \textbf{R}, and \textbf{P}, respectively. Moreover, by comprehensively evaluating the performance across all target domains, our method significantly outperforms other compared approaches in overall cross-domain generalization capability, further validating its effectiveness.

\subsection{Scalability to Image Classification}
Although our method is originally designed for domain generalization object detection, it can also be effectively transferred to image classification tasks to improve the domain generalization performance of classification models. \tabref{tab:classification} presents the multi-source domain generalization image classification results on the PACS dataset~\cite{li2017deeper}, where our method was applied to an image classification model with ResNet-50 as the backbone network. The results indicate that our proposed GB-DAL method successfully enhances the domain generalization performance of DAL-based classification models, improving accuracy from 79.6\% to 80.5\%. Furthermore, with the additional integration of the SNF module, the model's average performance is further promoted to 83.2\%, fully validating the effectiveness and generalizability of the proposed method in image classification tasks. To further validate the effectiveness of our SNF, we incorporate it into 4 typical DG models, i.e., FACT\cite{xu2021fourier}, FSDR~\cite{huang2021fsdr}, MADG \cite{dayal2023madg} and RASP \cite{kim2024randomized} for image classification. The experiments demonstrate that, although our method is originally designed to enhance the domain generalization object detection performance, it can also be effectively transferred to image classification tasks, significantly improving the domain generalization performance of classification models.

\begin{table}[tb]
	\begin{center}
			\begin{tabular}{l|c c c c c}
				\hline
				Method               & \textbf{P}      & \textbf{A}   & \textbf{C}      & \textbf{S}    & \textbf{Avg}\\
				\hline
				DAL                  & 91.9            & 76.9         & 75.8            & 73.7          & 79.6          \\
				GB-DAL                 & 95.2            & 79.9         & 76.4            & 70.4          & 80.5          \\
				+\textbf{SNF}             & \textbf{95.5}   & \textbf{82.1}& \textbf{77.0}   & \textbf{78.0}  & \textbf{83.2}  \\
                \hline
				FACT\cite{xu2021fourier}             & 95.2   & 85.4& 78.4   & 79.2  & 84.5  \\
				+\textbf{SNF}             & \textbf{96.3}   & \textbf{85.6}& \textbf{78.9}   & \textbf{80.2}  & \textbf{85.2}  \\
                \hline
				FSDR \cite{huang2021fsdr}             & 96.3   & 87.2 & 80.1   & 79.4  & 85.8  \\
				+\textbf{SNF}             & \textbf{97.3}   & \textbf{88.2}& \textbf{80.2}   & \textbf{80.6}  & \textbf{86.5}  \\
              \hline
				MADG \cite{dayal2023madg}            & 96.5   & 88.2 & 80.2   & 80.1  & 86.3  \\
				+\textbf{SNF}            & \textbf{97.3}   & \textbf{88.6} & \textbf{81.6}   & \textbf{81.3}  & \textbf{87.2}  \\
                \hline
				RASP \cite{kim2024randomized}             & 94.1   &84.6 & 79.8   & 80.1  & 84.7  \\
				+\textbf{SNF}             & \textbf{95.2}   & \textbf{85.2}& \textbf{80.7}   & \textbf{80.5}  & \textbf{85.4} \\
				\hline
			\end{tabular}
    \end{center}
    \caption{Results of multi-source DG for image classification, trained on three domains and tested on the rest domain in PACS \cite{li2017deeper}. 4 recent DG models including FACT~\cite{xu2021fourier}, FSDR~\cite{huang2021fsdr}, MADG \cite{dayal2023madg} and RASP \cite{kim2024randomized} for image classification are conducted to validate the effect of SNF.}
    \label{tab:classification}
\end{table}

\begin{table}[tb]
\begin{center}
		\setlength{\tabcolsep}{1mm}{
			\begin{tabular}{ c c c |c c c }
				\hline
				\multicolumn{3}{c|}{Method} &   \multicolumn{3}{c}{mAP} \\
				GB-DAL$_{local}$ & GB-DAL$_{global}$ & SNF  & to F         &  to C           & to B    \\
				\hline
				&  &    & 31.5          & 37.2            & 24.8         \\
				
				\checkmark  &               &                & 34.1          & 41.4            & 25.6         \\
				\checkmark  & \checkmark    &                 & 34.8          & 42.3            & 26.4         \\
				&   & \checkmark     & 35.6          & 43.4            & 25.8         \\

			  \checkmark  & \checkmark    & \checkmark               & 39.1         & 44.9            & 27.7         \\
				\hline
		\end{tabular}}
	\end{center}
	\caption{Ablation results by training on two domains of \{\textbf{C}, \textbf{F}, \textbf{B}\} and tested on the rest one. The 1st row is the result of the vanilla Faster RCNN\cite{ren2015faster}.
	}
	\label{tab:ablation1}
\end{table}
		
\vspace{0.5cm}		
\subsection{Ablation Study}
\textbf{Effectiveness of different components.}
To validate the effectiveness of each component in the proposed method, we design  ablation experiments on the \textbf{F}, \textbf{C}, and \textbf{B} datasets. Specifically, in each experiment, two datasets are used as source domains, while the remaining dataset serves as the target domain.
Three components are discussed, including GB-DAL$_{local}$, GB-DAL$_{global}$ and SNF, by adding them sequentially and observe the change in detection performance.
As shown in \tabref{tab:ablation1}, the ablation results demonstrate that each component of our proposed method contributes positively to model performance across different settings. 
Specifically, the performance of the vanilla Faster RCNN \cite{ren2015faster} is improved progressively with the successive addition of local GB-DAL and global GB-DAL (+2.6\% and +0.7\% on task \textbf{C}\&\textbf{B}$\to$\textbf{F}), indicating that both local and global GB-DAL effectively mitigate intra-dataset non-causal factors at their respective levels, thereby enhancing the detector's generalization to unseen domains. 
As a plug-and-play data augmentation module, SNF also demonstrates performance gains of 4.3\% and 4.1\% on task \textbf{C}\&\textbf{B}$\to$\textbf{F} by  training with GB-DAL based Faster RCNN and vanilla Faster RCNN, respectively. This confirms that SNF reduces dataset bias by injecting simulated non-causal factors, thereby promoting causal invariance. 
\textbf{Robustness to adversarial and natural noise}. 
We find that the proposed SNF not only achieves our initial goal of enhancing model generalization but also exhibits a certain degree of robustness against adversarial noise and natural noise. 
As shown in \tabref{tab:adv_gauss}, compared to the GB-DAL method, introducing the SNF module during training significantly improves model performance on the clean dataset \textbf{F}, the adversarial perturbation dataset $\mathrm{\textbf{F}_{\textbf{adv}}}$, and the Gaussian (natural) noise dataset $\mathrm{\textbf{F}_{\textbf{gauss}}}$ as the unseen domain, with performance gains of 4.3\%, 7.7\%, and 9.8\%, respectively. These results demonstrate that the proposed SNF module not only effectively enhances the model's generalization ability to unseen domains but also substantially improves its robustness against adversarial attacks and natural noise.


\begin{table}[tb]
	\centering
		\begin{tabular}{ccccc}
			\cline{1-4}
			\multicolumn{1}{l|}{Method}         & \textbf{F}                    & $\mathrm{\textbf{F}_{\textbf{adv}}}$               & $\mathrm{\textbf{F}_{\textbf{gauss}}}$            &  \\ \cline{1-4}
			\multicolumn{1}{l|}{DAL}     & 27.0                 & 13.5                   & 9.6                 &  \\
			\multicolumn{1}{l|}{\textbf{GB-DAL}}     & 34.8                 & 19.0                   & 13.8                 &  \\
            
			\multicolumn{1}{l|}{\textbf{GB-DAL}+\textbf{SNF}} & \textbf{39.1}                 & \textbf{26.7}                 & \textbf{23.6}                 &  \\ \cline{1-4}
			\multicolumn{1}{l}{}          & \multicolumn{1}{l}{} & \multicolumn{1}{l}{} & \multicolumn{1}{l}{} & 
		\end{tabular}
		\caption{Results on clean dataset \textbf{F}, adversarial disturbance dataset $\mathrm{\textbf{F}_{\textbf{adv}}}$, and Gaussian noise dataset $\mathrm{\textbf{F}_{\textbf{gauss}}}$. The model trained on \textbf{C} and \textbf{B}
		}
		\label{tab:adv_gauss}
	\end{table}

\textbf{Dual mechanism of SNF.} 
In the SNF module, adversarial perturbations are introduced to simulate non-causal factors, thereby mitigating data bias and reducing the implicitness of non-causal factors. Meanwhile, adversarial samples incorporating these perturbations are also fed into the detection head for supervised training, essentially serving as a data augmentation strategy. Although previous experiments have demonstrated the effectiveness of SNF in enhancing model generalization, it is still necessary to conduct an ablation study to verify whether both mechanisms of adversarial perturbation—namely, non-causal factor simulation and data augmentation—contribute to improving generalization performance. As shown in \tabref{tab:ablation_pert}, we progressively incorporate the two mechanisms of adversarial perturbation into the GB-DAL based object detection framework. Notably, for non-causal factors simulation mechanism (\ie, ``Sim''), the adversarial samples are only used for GB-DAL (domain alignment) without feeding into the detection head; for data augmentation mechanism (\ie, ``Aug''), the adversarial samples attend the training of the detection head. The experimental results indicate that both non-causal factor simulation and data augmentation significantly improve model performance on unseen target domains. Besides, we conduct a comparative experiment of conventional image augmentation (i.e., IA with flipping, rotation and color jittering), adversarial perturbation-based gradient augmentation approaches (i.e., GA\cite{huang2022adversarial}) and data mixing augmentation  (i.e., AdAutomixup \cite{qin2024adversarial}). The results show that SNF is superior to IA, GA and AdAutomixup. This finding fully validates the effectiveness of the dual mechanism of adversarial perturbation and its positive contribution to model generalization performance.

\begin{table}[tb]
	\centering
		\begin{tabular}{l|c c c c c }
			\hline
			Method  & C\&B→F  & F\&B→C  & C\&F→B   \\
			\hline
			GB-DAL  &  34.8  & 42.3   & 26.4   \\
             GB-DAL+IA  & 35.2  & 42.7   & 26.6   \\
			GB-DAL+GA    &  37.6   & 43.4    & 26.9  \\
            GB-DAL+AdAutoMix    & 38.2   & 44.2    & 27.0  \\
            
			GB-DAL+Sim    &  37.6   &  44.1    & 27.2  \\
        
			GB-DAL+Sim+Aug  &   39.1  & 44.9  & 27.7  \\
			\hline
		\end{tabular}
\caption{Ablation results on the effectiveness of the dual mechanism of adversarial perturbation from SNF. ``IA'', ``GA'',``AdAutoMix''  and ``Sim'' represent conventional image augmentation (e.g. flipping, rotation, etc.), adversarial perturbation-based gradient augmentation approach in \cite{huang2022adversarial}, data mixing augmentation in \cite{qin2024adversarial} and the non-causal factor simulation method, respectively. ``Sim+Aug'' adopts the dual mechanism of SNF and data augmentation during training.}
\label{tab:ablation_pert}
\end{table}

\begin{figure*}[t]
	\centering
	\includegraphics[width=1\linewidth]{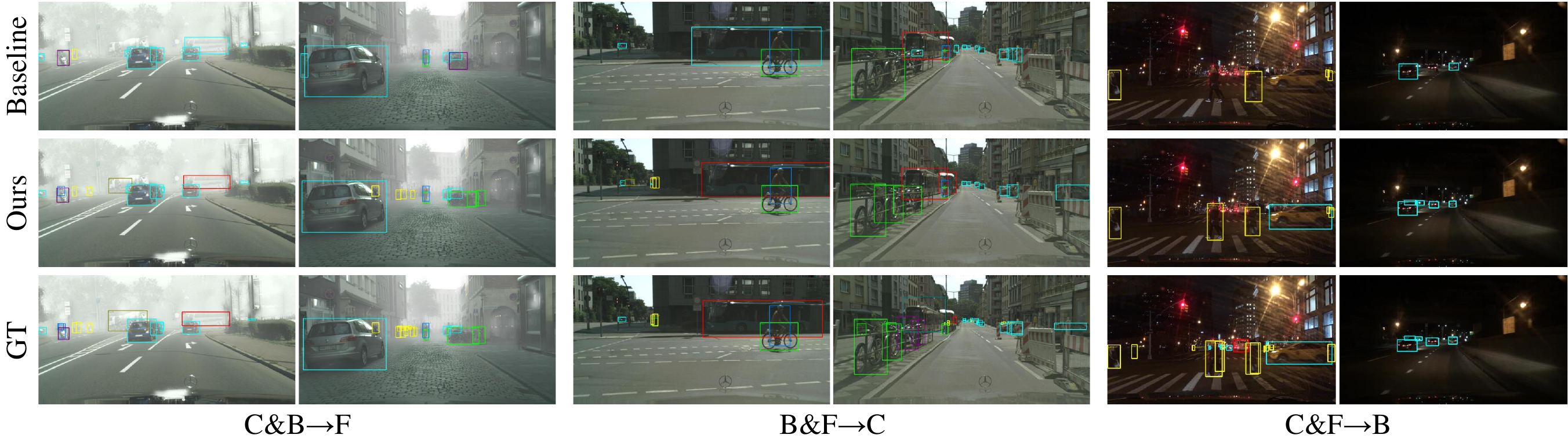}
	\caption{Qualitative comparison between the baseline and our proposed method. Notably, GT means the ground truth annotations. 
	}
	\label{fig:det_vis}
\end{figure*}

\textbf{Qualitative results.} \figref{fig:det_vis} presents the visualization of detection results on different target domains in three cross-domain settings. From the visualized results, it can be observed that our method consistently outperforms the baseline across all three experiments. Furthermore, our approach effectively mitigates the false positives and false negatives that the baseline always encounters in unknown target domains. This qualitative analysis further demonstrates the effectiveness of our proposed method in enhancing the model's generalization capability for object detection in unseen domains.

\subsection{Hyperparameter Analysis}

In this subsection, we investigate the impact of three key hyperparameters involved in the proposed GB-DAL and SNF methods on \textbf{C}\&\textbf{B}$\to$\textbf{F} task. These hyperparameters include the number \( K \) of domains in the GB-DAL method, the trade-off parameter \( \lambda \) in the overall training loss function of \equref{eq:loss_total}, and the adversarial perturbation coefficient \( \epsilon \) in the SNF module of \equref{eq:adversarial_pert}. Through experimental analysis, we systematically examine the influence of these parameters on model performance and determine their optimal values.

\begin{figure}[t]
	\centering
	\subfigure[$K$\label{fig:parameter_k}]{
		\includegraphics[width=0.37\linewidth]{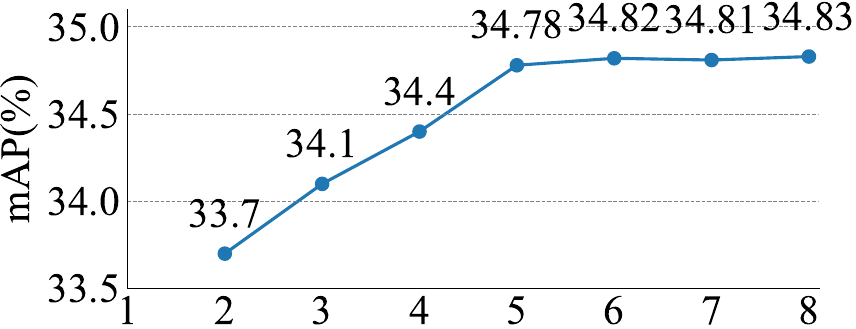}}
	\subfigure[$\lambda$\label{fig:parameter_lambda}]{
		\includegraphics[width=0.28\linewidth]{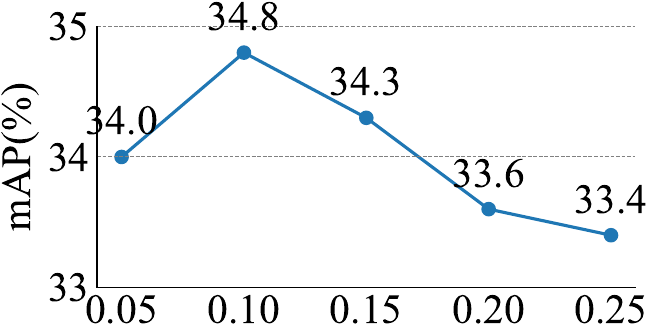}}
	\subfigure[$ \epsilon $\label{fig:parameter_eps}]{
		\includegraphics[width=0.28\linewidth]{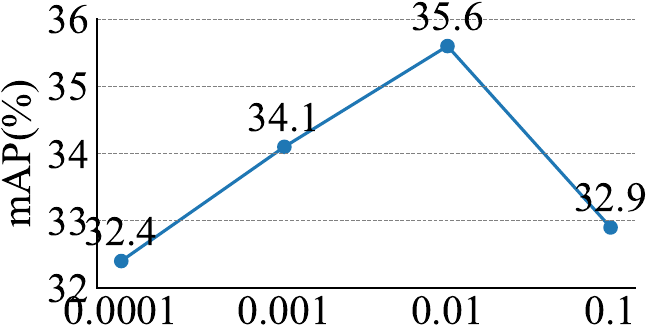}}
	\caption{Parameter sensitivity analysis on the task \textbf{C}\&\textbf{B}$\to$\textbf{F}. The parameters include the number \( K \) of domain splits, the trade-off parameter \( \lambda \) and the perturbation coefficient \( \epsilon \).
	}
	\label{fig:parameter}
\end{figure}

\textbf{The number $ K $ of domain splits}. 
In the GB-DAL method, the hyperparameter \( K \) determines the number of fine-grained domains via datasets splitting. This mechanism is used to first indicate and then filter implicit non-causal factors across different domains via DAL. Intuitively, a larger \( K \) allows for a more fine-grained identification of non-causal factors within a dataset. However, increasing \( K \) also introduces additional computational overhead. As illustrated in \figref{fig:parameter_k}, the performance is improved progressively with increasing \( K \) and reaches saturation at \( K=5 \). Thereby, we set \( K \) to 5 to achieve an trade-off balance between effectiveness and efficiency.

\textbf{Trade-off parameter $\lambda$}. 
The trade-off parameter \( \lambda \) in \equref{eq:loss_total} is used to balance the optimization of the detection head and the GB-DAL based alignment. To determine an optimal value, we conduct a grid search over the range of 0.05 to 0.25, selecting five candidate values for hyperparameter analysis. As shown in \figref{fig:parameter_lambda}, the model achieves the best generalization performance on the unseen target domain when \( \lambda = 0.1 \). Therefore, we set \( \lambda \) to 0.1 to achieve an optimal balance between detection head training and GB-DAL training.

\textbf{Perturbation coefficient $ \epsilon $}. 
The perturbation coefficient \( \epsilon \) in \equref{eq:adversarial_pert} controls the intensity of the adversarial perturbation. As shown in \figref{fig:parameter_eps}, the results indicate that both excessively large and excessively small values of \( \epsilon \) lead to a performance decline. This is due to that a large \( \epsilon \) introduces excessively strong image perturbations, potentially disrupting causal factors in the image. Conversely, a small \( \epsilon \) fails to effectively introduce non-causal factors to mitigate data bias, making it difficult to improve the model's generalization ability. Based on the experimental analysis, we set \( \epsilon = 0.01 \) to achieve an optimal generalization performance.

\section{Conclusion and Future Work}

This paper deeply delves into the non-causal factors implied in domain-invariant representations for most DG models, which are detrimental to open-world object detection. To uncover the veil of non-causal factors, two critical findings are observed. 1) DAL can only capture inter-dataset non-causal factors, but overlook many intra-dataset non-causal factors due to the dataset-level sparse domain label. 2) The implicitness of non-causal factors disable the model learning causal feature due to dataset bias. Based on the two findings, we contribute two new ingredients: GB-DAL and SNF. The former aims to furnish the domain discriminator with denser domain labels via datasets split under Granular-Ball computing perspective. The latter introduces simulated non-causal factors under an adversarial perturbation perspective to bridge data bias. Numerous experiments on domain generalization object detection (DGOD) show the superiority of our method. 

The proposed method brings two unexpected benefits. 1) It can work well for DG image classification task (\tabref{tab:classification}), since this is a plug-play model, although the proposed method is originally designed for object detection. 2) It can obtain adversarial robustness and natural robustness, brought by SNF, as a free-lunch (\tabref{tab:adv_gauss}), although this work pursuits generalization. Substantially, such robustness may come from an approximate adversarial training of the DG object detector. Therefore, this inspires us to raise an open question: \textit{how to simultaneously guarantee the model generalization and robustness in open-world vision tasks?} This remains a challenging problem worth exploring in future work.

\begin{small}
\vspace{.3in} \noindent \textbf{Data Availability.}
The data used in this study include six object detection benchmark datasets and one image classification benchmark dataset, i.e., Cityscapes [\href{https://www.cityscapes-dataset.com}{https://www.cityscapes-dataset.com}], Foggy Cityscapes [\href{https://www.cityscapes-dataset.com}{https://www.cityscapes-datas\linebreak et.com}], Rain Cityscapes [\href{https://www.cityscapes-dataset.com}{https://www.cityscapes-dataset.com}], SIM10k [\href{https://fcav.engin.umich.edu/projects/driving-in-the-matrix}{https://fcav.engin.umich.edu/projects/driving-in-the-matrix}], PASCAL VOC [\href{http://host.robots.ox.ac.uk/pascal/VOC/}{http://host.robots.ox.ac.uk/pascal/V\linebreak OC/}], BDD100k [\href{http://bdd-data.berkeley.edu/}{http://bdd-data.berkeley.edu/}], and PACS [\href{https://sketchx.eecs.qmul.ac.uk/}{https://sketchx.eecs.qmul.ac.uk/}].

%


%

\end{small}

\begin{acknowledgements}
This work was partially supported by National Natural Science Fund of China under Grants 92570110, 62271090 and 62221005, Chongqing Natural Science Fund under Grant CSTB2024NSCQ-JQX0038, National Key R\&D Program of China under Grant 2021YFB3100800 and National Youth Talent Project. 


\end{acknowledgements}

\bibliographystyle{unsrt}
\bibliography{reference}

\end{document}